\documentclass[11pt,a4paper]{article}
\usepackage[hyperref]{acl2021}
\usepackage{times}
\usepackage{latexsym}
\usepackage{amsmath}

\usepackage{graphicx}%,dblfloatfix}
\usepackage{amssymb}
\usepackage{subcaption}
\usepackage{array,booktabs,multicol}
\usepackage{microtype}
\usepackage{afterpage}

\aclfinalcopy % Uncomment this line for the final submission
\usepackage{xspace}
\usepackage{multirow}

\newcommand{\npone}{S\xspace}
\newcommand{\nptwo}{O\xspace}
\newcommand{\roberta}{{\sc r}o{\sc bert}a-{\sc l}\xspace}
\newcommand{\bert}{{\sc bert}\xspace}

\newcommand{\electra}{{\sc electra}\xspace}

\newcommand{\electralarge}{{\sc electra-l}\xspace}
\newcommand{\gpt}{{\sc gpt}\xspace}
\newcommand{\gpttwo}{{\sc gpt2-m}\xspace}
\newcommand{\xlm}{{\sc xlm}-{\sc r}\xspace}
\newcommand{\mbert}{m{\sc bert}\xspace}

\title{John praised Mary because \textit{he}?\\
Implicit Causality Bias and Its Interaction with Explicit Cues in LMs}
\author{Yova Kementchedjhieva \\
  University of Copenhagen\\
  \texttt{yova@di.ku.dk} \\\And
  Mark Anderson \\
  Universidade da Coru\~na, CITIC\\
 Department of CS \& IT \\
  \texttt{m.anderson@udc.es} \And
  Anders S{\o}gaard \\  
  University of Copenhagen\\
  \texttt{soegaard@di.ku.dk} \\}

\date{}

\begin{document}
\maketitle
\begin{abstract}
%I'd suggest changing the focus of the abstract. Don't explain the phenomenon. Say what you did. 

Some interpersonal verbs can implicitly attribute causality to either their subject or their object and are therefore said to carry an implicit causality (IC) bias. Through this bias, causal links can be inferred from a narrative, aiding language comprehension. We investigate whether pre-trained language models (PLMs) encode IC bias and use it at inference time. We find that to be the case, albeit to different degrees, for three distinct PLM architectures. However, causes do not always need to be implicit---when a cause is explicitly stated in a subordinate clause, an incongruent IC bias associated with the verb in the main clause leads to a delay in human processing. We hypothesize that the temporary challenge humans face in integrating the two contradicting signals, one from the lexical semantics of the verb, one from the sentence-level semantics, would be reflected in higher error rates for models on tasks dependent on causal links. The results of our study lend support to this hypothesis, suggesting that PLMs tend to prioritize lexical patterns over higher-order signals. 

%We find that implicit causality bias is reliably captured by all models considered, with bidirectional encoding seemingly a prerequisite for the more effective acquisition of this semantic property. We also find consistent interference effects from implicit causality in the context of a contradicting explicit cause, with a larger, more richly trained model proving least susceptible to this effect.

\end{abstract}

\section{Introduction}

Recognising causal links in narrative is an integral component of language comprehension that often relies on implicit cues \cite{trabasso1985causal}. Pre-trained language models, which form the basis of many language processing solutions nowadays, should therefore pick up on such cues and integrate them correctly with other signals to enable accurate causal inferences in downstream tasks, including question answering and information extraction. 
%Some interpersonal transitive verbs trigger causal attribution of the action or change of state that they denote to their subjects or objects. 
%Consider as an example the sentence `John loves Tim because he is so \textit{flod}'---without having any knowledge of the unknown (and madeup) word \textit{flod}, most people would agree that Tim is the more plausible antecedent to the ambiguous pronoun \textit{he}, and that \textit{flod} likely describes a property of his that makes him lovable.
%rather than a property of John that makes him loving. 

\begin{figure}
    \centering
    \includegraphics[width=\linewidth]{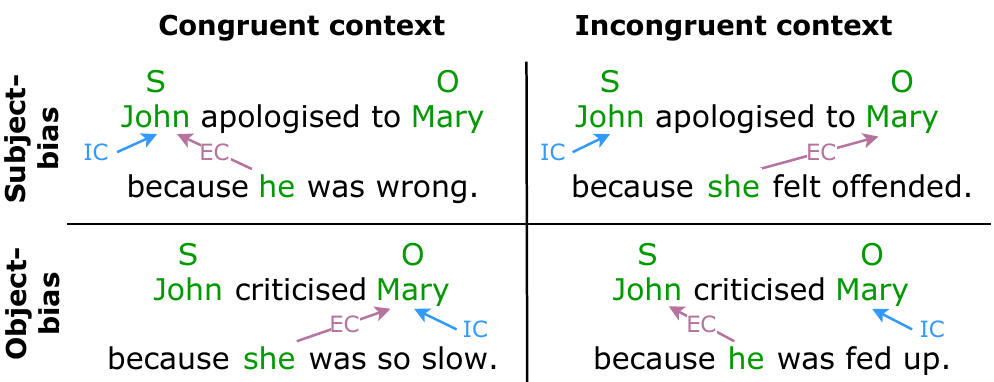}
    \caption{Illustration of implicit causality (IC) and explicit causality (EC) in contexts where the two are congruent, and where they are incongruent. 
    %Implicit causality is expressed by lexical semantics and independent of context: \textit{apologise} implicates causality on its subject; \textit{criticise} implicates causality on its object. Explicit causality is expressed by sentential semantics, so it is contextual. Explicit causality can be either congruent or incongruent with the verb's implicit bias, i.e. it can attribute cause to the same participant as the verb implicitly does, or to the other participant, respectively.
    }
    \label{fig:task}
\end{figure} 

Psycholinguists have identified one such cue in the implicit causality bias of interpersonal verbs: some interpersonal verbs %of this type 
tend to implicate causality on either their subject or their object \cite{10.2307/4177835}. It is this bias that leads to wide agreement among subjects in psycholinguistic studies, that in a sentence like \textit{John appreciates Mary}, the cause for appreciation likely lies with a property or action of Mary's rather than John's. Causality can also be stated explicitly, in the form of a subordinate \textit{because} clause, for instance, and it can optionally contradict the implicit causality of the verb in the main clause. In Figure~\ref{fig:task}, we show two verbs in the context of explicit statements of causality (EC) that are either congruent or incongruent with their IC bias. Psycholinguistic studies show that congruency affects language comprehension, with human participants taking longer to identify the referent to the pronoun after \textit{because} in incongruent contexts compared to congruent ones \cite{caramazza1977comprehension}. The integration of the two signals costs humans extra effort, but they are eventually able to overcome the false initial expectation based on lexical semantics (i.e. IC) and form a final response that takes into consideration the full sentence-level semantics (i.e. EC). \citet{ettinger2020bert} suggests that in the context of such diverging signals, models would likely fail to integrate all signals correctly, producing a response that is consistent with the initial, shallow expectation and therefore incorrect.

In this work, we study a range of large transformer-based pre-trained language models (PLMs) with a focus on their awareness of IC bias and their response to stimuli of the kind shown in Figure~\ref{fig:task}. Following \citet{ettinger2020bert}, we hypothesize that a language model aware of IC bias would experience interference from this signal in the context of incongruent EC, resulting in errors of judgement on a co-reference resolution task dependent on causal inference.

In a study of six PLMs from three model families: unidirectional generative, bidirectional generative and bidirectional discriminative, we find that IC bias is reliably encoded by all, but not used to an equal degree when making predictions in a controlled setting designed to test for IC awareness. In line with our hypothesis, we find that models with high IC awareness suffer an interference from IC bias in contexts of incongruent causality. We discuss these findings with reference to model type, size and amount of training data; we also draw general conclusions about the shortcomings of language models, which seem to prioritise a low-level lexical pattern (when they are aware of it in the first place) over a higher-order contextual signal.

%While this has been previously done for unidirectional autoregressive PLMs \cite{davis-van-schijndel-2020-discourse, upadhye-etal-2020-predicting}, most innovations in the space of language modeling at the moment use bidirectional encoding. We therefore introduce a procedure that allows us to also test bidirectional PLMs, both masked and discriminative. In a comparison of four competitive PLM architectures, we find that bidirectional encoding supports better learning of IC bias.  Lack of congruency between these two signals is indeed found to interfere with the models' ability to resolve co-reference, although a larger, more richly trained model is clearly less susceptible to this effect than others. 

\section{Related Work}

The study of the linguistic capacities of neural language models (LMs) has become especially relevant in current NLP research,
where representations from PLMs feed into systems for various complex tasks, typically improving performance.  %with promising results. 
Many of the testing paradigms used in psycholinguistics lend themselves well to LM analysis as they rely on a textual stimulus and a lexical response.

\citet{linzen2016assessing} were first to borrow from the psycholinguistic testing paradigm, in a study of the capabilities of LSTM-based models to resolve subject-verb number agreement. %---that the language modeling objective alone was not sufficiently sensitive to this phenomenon. \citet{giulianelli} focused on the same phenomenon in an investigation of where and when subject and verb agreement is encoded in LMs. 
%In another psycholinguistics-inspired study, \citet{futrell2019neural} assessed the ability of different neural LMs to represent syntactic states, finding that various RNN-based models learn basic syntactic state representations, but a large size is a prerequisite for the learning of finer-grained details.
\citet{goldberg2019assessing} adopted the psycholinguistic approach in an assessment of \bert \cite{devlin2018bert} on a number of syntactic tasks and found it to perform remarkably well on all. \citet{hawkins-etal-2020-investigating} studied the ability of  different LMs to capture human preferences as to the argument structure of English verbs.%, a phenomenon at the interface of syntax and semantics.

The analysis of semantic capabilities in LMs includes studies on negative polarity in LSTM LMs \cite{marvin2018targeted, jumelet2018do}, reasoning based on higher-order linguistic skill  \cite{talmor2019olmpics}, arithmetic and compositional semantics \cite{staliunaite2020compositional}, stereotypic tacit assumptions and lexical priming \cite{misra2020exploring,weir2020existence}. 
Many of these studies look at recent PLMs and draw mixed conclusions about the level of semantics encoded by these models. \citet{peters2018dissecting} and \citet{tenney2019bert} observed that PLMs do encode some higher-order syntactic abstractions in the higher layers (whereas lower-order syntactic information is encoded in the lower layers). However, in a comparison of contextualized and static word embeddings, \citet{tenney2019you} concluded that PLMs do not generally offer the same improvement with respect to semantics as they do for syntax.  

At the crossroad of semantic analysis and psycholinguistic approaches, \citet{ettinger2020bert} introduced a suite of six psycholinguistic diagnostics for the analysis of semantic awareness in LMs. %, and applied it in an evaluation of \bert. 
The tasks were selected based on a specific pattern observed in the response of human participants in psycholinguistic studies: an initial expectation (marked by an N400 electrophysiological response) that diverges from the final answer in a cloze task that humans offer once they have had time to fully consider the test prompt. \citet{ettinger2020bert} suggests that LMs might be ``tripped up'' in such contexts if they are unable to accurately integrate all the available information---she indeed found that to be the case for role-based event prediction in \bert \cite{devlin2018bert}, for example. The phenomenon we study, incongruency in causality signals, has been observed to trigger a similar response in humans \cite{van2007establishing} and can thus be expected to also ``trip up'' LMs. 

\paragraph{Implicit causality bias} was previously considered in PLM analysis by two works, both looking at how well unidirectional PLMs capture it. \citet{upadhye-etal-2020-predicting} studied IC from the perspective of how different connectives between the main clause and the following clause (\textit{because}, \textit{and as a result}, full stop) affect the strength of the bias. While they did not find strong evidence for a correlation to human-based results in this respect, they did observe that in the context of connective \textit{because} PLMs assigned lower probability to subject-referring pronouns for an object-biasing verb as compared to a subject-biasing verb. 
\citet{davis-van-schijndel-2020-discourse} observed that GPT2-XL \cite{radford2019language} encodes some level of IC bias in its representations (measured in terms of similarity between the representation of the pronoun and its two potential referents) and its decision on how to resolve a referent at prediction time is weakly influenced by that. They took the analysis one step further and looked at whether GPT2-XL uses IC information to resolve relative clause attachment, which in humans is conditioned by IC bias---no evidence was found to suggest that that was the case. Our study extends previous work on IC bias in several ways: we study both unidirectional and bidirectional models, we measure bias in the same terms as was done in psycholinguistic work and can therefore assess the correlation between the two, and we study the matter from a perspective that has not been considered before, namely the case of incongruent explicit and implicit causality. %This allows us to draw conclusions not just about lexical semantics but also about its integration with sentence-level semantics.  

\section{Materials}
\label{sec:materials}
Here we describe the two psycholinguistic diagnostics that we draw on: for the study of IC bias in isolation and its integration with EC. We also describe the modifications necessary to make these diagnostics suitable for PLMs.

%\subsection{Context-free bias effects}
\subsection{Context-free IC Bias}
\label{subsec:exp1}

\citet{ferstl2011implicit} studied IC bias in a context free of EC through a sentence completion task where subjects were presented with a stimulus like 
\begin{enumerate}
   \item[(1)]\label{itm:incon1} John praised Mary because \underline{\hspace{0.5cm}} 
\end{enumerate}

\noindent and asked to finish the sentence. Continuations were observed to start with a third person pronoun (\textit{he} or \textit{she}) 94.2\% of the time. The researchers counted the ratio of continuations referring back to the subject of the sentence, $s_{wins}$, and back to the object, $o_{wins}$, and computed a bias score for each verb as $100\times(s_{wins}-o_{wins})/(s_{wins}+o_{wins})$. This results in a range of $-100$ (verbs with extreme Object bias, hereafter \nptwo-bias)  to $100$ (verbs with extreme Subject bias, hereafter \npone-bias). %The authors set the threshold for significant bias at +/-26. %The materials used in the study and the results obtained are publicly available. 
The study covered 305 interpersonal English verbs with responses from 96 subjects. %---both made available by the authors.

In another study of IC bias, \citet{hartshorne2013verb} presented subjects with stimuli with a nonce ending, e.g.

\begin{enumerate}
   \item[(2)]\label{itm:incon2} John praised Tim because he was a \textit{dax}. 
\end{enumerate}

\noindent and asked them the question \textit{Who do you think is a dax?} The nonce ending is meant to provide a
content-free continuation that does not affect the interpretation of the ambiguous pronoun, neither semantically (as the madeup \textit{dax} carries no meaning), nor syntactically---\citet{hartshorne2015causes} conclude that explanations of the form \textit{is a/an X} do not affect people’s intuitions about who the explanation referred to. 

\paragraph{Our approach} is to use the procedure of \citet{ferstl2011implicit} as is to test unidirectional PLMs, as they are naturally suited to the open-end input format. Since bidirectional PLMs have been trained on complete utterances and may thus act unpredictably in an open-end context,\footnote{We find that a common response of generative bidirectional PLMs to stimuli like those in (1) is to predict a full-stop for the empty slot.} we test such models with a modification of the procedure of \citet{hartshorne2013verb}:\footnote{See Appendix~\ref{app:alttask} for an alternative we considered.} we convert it into a cloze task with a gender mismatch between the participants, such that (3) becomes 

\begin{enumerate}
   \item[(3)]\label{itm:incon3} John praised Mary because \underline{\hspace{0.5cm}} was a \textit{dax}. 
\end{enumerate}

We adopt the mismatched-gender setting as it more closely resembles the sentence completion task in \citet{ferstl2011implicit}. In both formats we can now identify the preferred referent by looking at the probability of pronouns \textit{he} and \textit{she} for the empty slot, each one referring unambiguously to only one referent. Inducing a prediction for a pronoun in the empty slot is also a more natural choice of co-reference than repeating one of the names \cite{holtzman2019curious}.

In the examples shown throughout the paper, \textit{John} and \textit{Mary} are used as placeholders for the subject and object of the verb of interest. The choice of names to go in these slots can affect model predictions \cite{abdou-etal-2020-sensitivity}, so we generate 200 variants of each stimulus, varying the names and the order between the two genders and we query the PLMs with all of them. The full procedure is described in Appendix~\ref{app:proper}. 

We compile a list of 200 nonce words using the 194 nonce words made available by \citet{bangert2012reaching}, five nonce words from \citet{CUSKLEY2015205}, manually chosen to resemble English nouns, and \textit{dax}, used in \citet{hartshorne2013verb}. When presenting a bidirectional PLM with the aforementioned 200 variants of a stimulus, we dynamically draw a nonce word at random from this list without replacement. 

The procedure described above is applied to each of the 305 verbs studied in \citet{ferstl2011implicit}.

%Analogous resources are available for Spanish \cite{goikoetxea2008normative} and German %\cite{van2018discourse},
%\footnote{We obtained the German data in personal communication with the first Emiel van den Hoven.} each consisting of 100 verbs. 
%\citet{ferstl2011implicit} reports that the correlation between the IC bias for the 42 verbs included in their study and that of \citet{goikoetxea2008normative} (translated to English for the comparison) is 0.69.   

%\subsection{Context-bound bias effects}
\subsection{IC Bias in the Context of EC}
\label{subsec:exp2}

\citet{caramazza1977comprehension} tested the effect of incongruency between IC and EC using pairs of sentences built around the same verb, where one contains an explanation congruent with the verb's bias and the other contains an incongruent explanation (refer back to Figure~\ref{fig:task} for some examples). Participants were shown one sentence at a time on a screen and asked to say out loud who the referent was to the pronoun after \textit{because}. %---in the examples above the choice would be between \textit{John} and \textit{Mary}. 
\citet{caramazza1977comprehension} carried out experiments both with stimuli where the referents are of the opposite gender and where the referents are of the same gender---responses were delayed in the context of incongruent explanations as compared to congruent ones in both settings, the effect being stronger in the mismatched-gender setting. 

\paragraph{Our approach} is to adopt the mismatched-gender setting and to convert this task into a cloze task as well, an example stimulus being:

\begin{enumerate}
   \item[(4)]\label{itm:incon4} John praised Mary because \underline{\hspace{0.5cm}} had done well.
\end{enumerate}

The stimuli used by \citet{caramazza1977comprehension} and other related studies like \citet{garnham1996locus} use only a handful of verbs (14 and 22, respectively). We therefore found it necessary to develop a more expansive dataset for the purposes of our study. Following the procedure described in Appendix~\ref{app:materials} we constructed pairs of subject-referring explanations and object-referring explanations for 99 verbs, 33 strongly subject-biased verbs (bias$>65$), 33 strongly object-biased verbs (bias$<-65$), and 33 verbs from the middle of the scale, which can be thought of as having no effect on the attribution of implicit cause. Selecting the verbs in this fashion, with large gaps between each group, allows us to see the difference between them most clearly. 

Similarly to before, 200 variants of each stimulus are generated, varying the names of the referents and the order between the two genders.

\section{Procedure}
\label{sec:procedure}
% [Basically should say:
% we study 6 models
% 3 of them chosen to be comparable in size/tr data but different in architecture/tr objective]
% The other 3 chosen to be 'bigger siblings' to those ones

% Inducing a prediction with a generative model is straightforward, so all good for the bi- gen- ones, but with the uni- ones we need a hack for the second task

% We use a similar hack for the discriminative models.

In this section we describe how we induce responses to the tasks described in \S\ref{sec:materials} for the two experiments in this study: measuring context-free IC bias and IC bias in the context of EC. Six English PLMs are considered in this study, representative of the unidirectional generative, bidirectional generative and bidirectional discriminative paradigms in language modeling. As seen in Table~\ref{tab:models}, \gpt, \bert and \electra are comparable in size and training data. \gpttwo, \roberta and \electralarge are included as the `bigger siblings' to the former three models respectively, selected to resemble closely the architecture of their counterparts, while having a larger size and richer training data.\footnote{For implementational details see Appendix \ref{app:models}.} Comparisons can therefore be made across the three base models, on one hand, and within each pair of a base model and its larger counterpart, on the other. The three larger models are comparable in size, but not fully comparable in training data, \gpttwo being trained on only a quarter of what the other two models are trained on. In a small multilingual experiment, we also experiment with German \bert, Spanish \bert and \mbert,\footnote{The size of the training data for mBERT is not exactly known---it consists of the 100 biggest Wikipedias. English Wikipedia, as made available on HuggingFace in 2019 is 14 GB in size; with the next 99 Wikipedias being 13\% the size of the English Wikipedia on average (based on number of articles), that works out to an estimate of 194GB in total.} a multilingual version of \bert.

\begin{table}[t!]
    \centering
    \resizebox{\linewidth}{!}{
    \begin{tabular}{rrrrrr}
    \toprule
     & Work & Size & Data& Dir & Obj\\
        \midrule
        \multicolumn{1}{l}{English}\\
        \cmidrule{1-1}
         \gpt & \citet{radford2018improving} & 110 & 16 & Uni & Gen\\
         \gpttwo & \citet{radford2019language} & 345 &40 & Uni & Gen\\
         \bert & \citet{devlin2018bert} & 110 & 16 & Bi & Gen\\
         \roberta & \citet{liu2019roberta} & 355 & 160 & Bi & Gen\\
         \electra & \citet{clark2019electra}& 110 & 16 & Bi & Disc\\
         %{\sc -small} & ---\texttt{"}--- & 14 & 16 & Bi & Disc\\
         \electralarge & ---\texttt{"}--- & 335 & 160 & Bi & Disc\\
         \midrule
         German \bert  & - & 110 & 12 & Bi & Gen\\
         %\gpttwo & - &  117 & 16 & Uni & Gen\\
         %\electra & - & 110 & 73 & Bi & Disc\\
         Spanish \bert  & \citet{CaneteCFP2020} & 110 & 20 & Bi & Gen\\
         \mbert &-& 110 & 194 & Bi & Gen\\
         %{\sc xlm-r} & \citet{conneau2019unsupervised} & 270 & 2,5K & Bi & Gen\\

    \bottomrule
    \end{tabular}
    }
    \caption{Model properties in terms of size (number of parameters in millions), training data (size in GB), directionality (uni- or bi-directional), and token-level training objective (generative or discriminative). 
    German \bert and \mbert are not the product of any published work, but are closely associated with \citet{devlin2018bert}.
    }
    \label{tab:models}
\end{table}

We first describe the procedure for bidirectional generative PLMs which is most straightforward. Both experimental tasks can be formulated as a cloze task (see Examples (3) and (4)). We place a mask tag in the empty slot, pass the input through the model and compute the probability that the models assigns to tokens \textit{he} and \textit{she} for the position of the mask tag.

The procedure is equally trivial when testing unidirectional PLMs for context-free IC bias effects, where the stimuli can naturally take on an open-ended form (see Example (1)). 
The partial sentence is passed through the model and a probability for the relevant pronouns is computed.
Measuring the effect of IC bias in the context of EC, on the other hand, cannot be performed in the next-word prediction paradigm, so for this task we instead use the unidirectional PLMs as language scorers: we create two versions of each stimulus, one with pronoun \textit{she}, one with pronoun \textit{he} in the empty slot, and obtain a probability for each as the average over the probabilities of all tokens in the sequence. 

The discriminative model \electra is trained to recognize replaced tokens in its input, i.e. for each token it computes a probability over two classes, \textit{replaced} ({\sc r}) and \textit{original} ({\sc o}). Based on the reasoning outlined in Appendix~\ref{app:electra} we conclude that the more appropriate way to probe \electra in our experiments is by taking the average over the probability of class {\sc o} for all tokens in a sequence, instead of looking at the probability of this class for the pronoun of interest alone. In a procedure similar to the one used for \gpttwo when used as a language scorer, we present \electra with two versions of each stimulus
, one with \textit{he}, one with \textit{she} in the slot of interest,
and we compute the average probability of class {\sc o} for each. We formalize the handling of all model types in Table~\ref{tab:inputs}. 

\begin{table}[t!]
    \centering
    \resizebox{\linewidth}{!}{
    \begin{tabular}{lllll}
    \toprule
      Model & Exp & p(he) =  \\
      \cmidrule{1-3}
      {\sc gpt(2)} & IC & p(he \textbar $w_1...w_4$)  \\
       & IC+EC &  p($w_1...w_4$ he $w_6...w_8$) \\
      ({\sc r}o){\sc bert}(a) &both&   p(\textsc{mask}==he \textbar $w_1...w_4$ \textsc{mask} $w_6...w_8$) \\ 
    \electra &both&p(class={\sc O} \textbar $w_1...w_4$ he $w_6...w_8$)  \\
    \bottomrule
    \end{tabular}
    }
    \caption{Probing PLMs for the effect of context-free IC (IC) and IC in the context of EC (IC+EC) experimental paradigm. $w_1...w_4$ = \textit{John praised Mary because}; $w_6...w_8$ = \textit{was a dax / had done well} for IC and IC+EC, resp. The procedure is analogous for \textit{she}.} 
    \label{tab:inputs}
\end{table}

\section{Exp. 1: Context-free IC Bias}

With this experiment we want to determine whether English PLMs exhibit the same tendencies as humans when it comes to the IC bias of actions/states expressed with interpersonal verbs in a context free of any explicit causes. To this end, we use the materials described in \S\ref{subsec:exp1} and the procedure described in \S\ref{sec:procedure}. For any given model, the IC bias per verb is measured over the responses of the model to the set of 200 stimuli, each response processed as follows. 
%: the stimulus is a partial sentence consisting of a subject and object with different grammatical gender, an interpersonal verb that relates these two entities, and the word \textit{because}, e.g. \textit{John apologised to Mary because}. The item of interest to us it is whether models prefer to see \textit{he} or \textit{she} in that context. 

\subsection{Measuring Bias}

For a sentence with a female subject and a male object, the probability of \textit{she} would be denoted as $p^{s}$, the probability of \textit{he} as $p^{o}$, and %a True value for
$p^{s}>p^{o}$ would indicate a preference for the subject for this
stimulus. %For a sentence with a male subject and female object, the opposite is true. 
Refer back to Table~\ref{tab:inputs} for a summary of how these probabilities are obtained with each model. 

Having obtained the values $p^{s}$ and $p^{o}$ for each of the 200 stimuli per verb, we next calculate the bias of this verb in the following manner: 

\begin{align}
    s_{wins} = \sum_{n \in N } (p_n^s - p_n^o) > 0 \\
    o_{wins} = \sum_{n \in N } (p_n^s - p_n^o) < 0 \\
    bias = 100\times\bigg(\dfrac{s_{wins} - o_{wins}}{s_{wins} + o_{wins}}\bigg)
\end{align}

where $N$ is the set of 200 stimulus variants per verb. 
This metric gives us a range from $-100$ (extreme \nptwo-bias) to $100$ (extreme \npone-bias), with 0 indicating an absence of any bias altogether.  

\begin{figure}[t!]
    \centering
    \includegraphics[width=\linewidth]{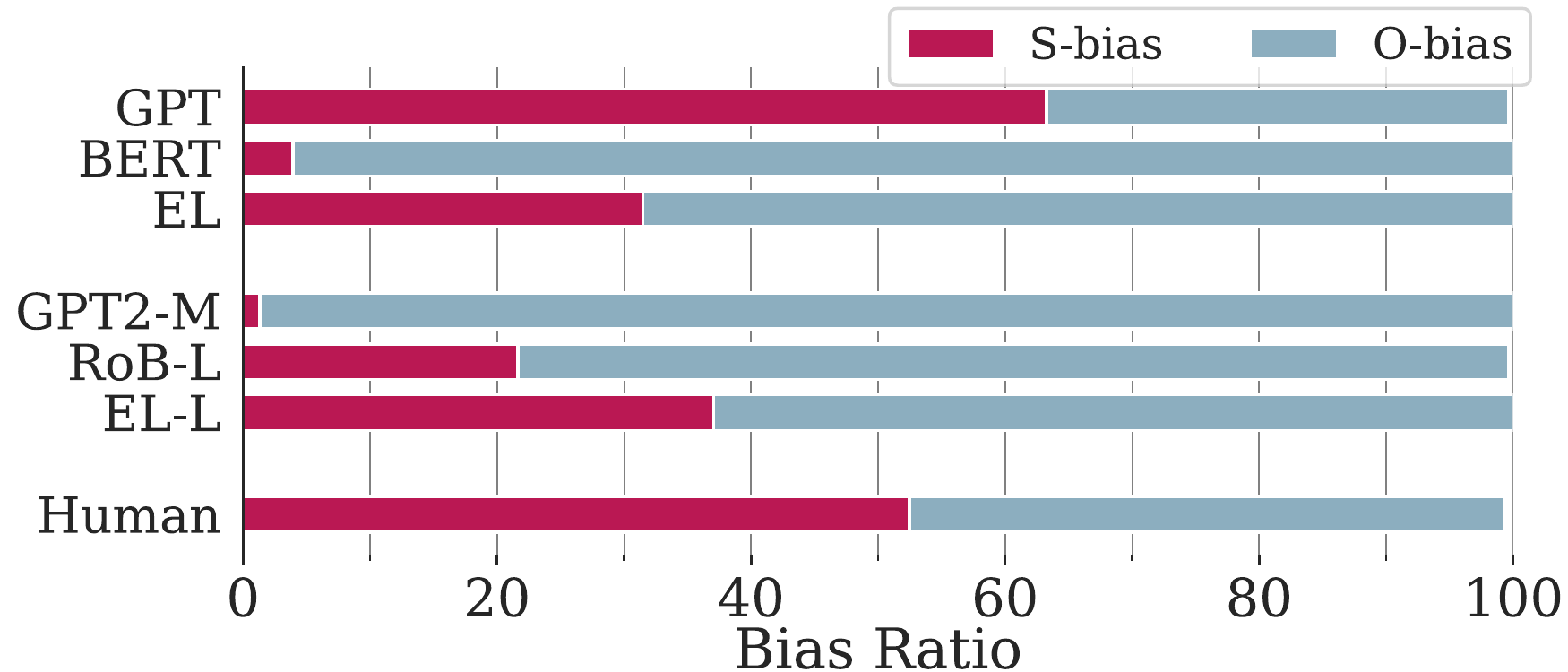}
    \caption{Ratio of S-bias verbs and O-bias verbs.}
    \label{fig:np1-ratios}
\end{figure}

\subsection{Preliminary analysis}
%As evidence that the procedure used in this experiment is valid,
As validation of the experimental procedure, we note that the generative models, \bert, \roberta, \gpt, \gpttwo, ranked one of the two vocabulary items of interest, % to us,
\textit{he} or \textit{she}, as their top prediction at a rate of 99.0, 99.4, 99.3 and 100.0 percent, respectively. It is reassuring to see that the models behave similarly to humans in this respect, who selected \textit{he} or \textit{she} at a rate of 94.2\% as a first token after \textit{because} \cite{ferstl2011implicit}. That also indicates that the probabilities assigned to the two tokens are meaningful \cite{holtzman2019curious}.

Figure~\ref{fig:np1-ratios} shows the ratio of S-bias verbs and O-bias verbs as determined by each of the models and by humans. Compared to human IC bias scores, which give an even distribution of verbs across the S-bias and O-bias classes,
%\footnote{\citet{ferstl2011implicit} report that the slight preference towards the subject-referring continuations was not found to be significant.} 
we see that most models show an imbalance in this respect--most notably, \bert and \gpttwo show an almost categorical preference for the object of the sentence. These trends could relate to the syntactic role of the participants (subject v. object), to their linear order, or to referent proximity. The first two factors are difficult to decouple in English, a language with a relatively fixed subject-verb-object word order. We discuss the effect of proximity in Appendix \ref{app:swap}.

To gain a clearer understanding of the IC bias awareness of the different models, we analyze the results of this experiment in their raw form and also with discounting for other potential sources of bias. In addition to the object bias discussed above, we include gender and choice of nonce words, which on their own did not appear to have a strong effect, but could combine with each other and with the object bias in unpredictable ways. The discounting for \textit{p(he)}, for example, in the context of stimulus \textit{John apologized to Mary because \_ was a dax} is done by subtracting the average probability of pronoun \textit{he} in the context of any stimulus with a male subject and word \textit{dax} in the nonce word slot (152.5 data points on average). 

\subsection{Results}
Table~\ref{tab:corr} quantifies the correspondence between model IC bias and human IC bias in terms of Spearman's $\rho$ over bias scores and in terms of micro-averaged F1 score over the polarity of the IC bias (subject-bias v. object-bias). For a plot of the exact bias values see Figure~\ref{fig:correlations-wugs} in Appendix \ref{app:vis}. The PLMs most affected by the discounting for other biases are \bert and \gpttwo, which also showed the strongest imbalance as observed in Figure~\ref{fig:np1-ratios}. All PLMs show a significant correlation to human IC bias, although this observation has the caveat of a small dataset (only 305 data points).

Within the pairs of related models, we can say that the differences between \bert and \roberta on the one hand and \electra and \electralarge on the other, are small, which suggests that already with 16GB of training data and 110M parameters, these architectures reach their potential in terms of capturing and using IC bias. For the two unidirectional PLMs, we see that after discounting \gpttwo exhibits a considerably higher correlation to human IC bias scores. This may indicate that the larger size and/or richer training data of \gpttwo have enabled the model to better capture IC bias, although the correlation still remains low in absolute terms.

\begin{table}[t!]
    \centering
    \resizebox{\linewidth}{!}{
    \begin{tabular}{l|llllll}
         \toprule
        & \bert & \textsc{r}o\textsc{b-l} & \gpt & \gpttwo & \textsc{el} & \textsc{el-l}\\
        \midrule
        $\rho$ & 0.58* & 0.67* & 0.22* & 0.22* & 0.72* & 0.72*\\
        F1 & 0.508 & 0.672 & 0.607 & 0.482 & 0.744 & 0.754\\
        \midrule
        $\rho$ & 0.65* & 0.69* & 0.23* & 0.38* & 0.73* & 0.71*\\
        F1 & 0.698 & 0.734 & 0.564 & 0.649 & 0.774 & 0.748\\
        \midrule
        \midrule
        LDA & 0.67* & 0.58* & 0.64* & 0.46* & 0.73* & 0.67*\\
        LR & 0.71* & 0.6* & 0.67* & 0.49* & 0.75* & 0.7*\\
    \bottomrule
    \end{tabular}
    }
    \caption{Correspondence between human- and model-induced IC bias scores (a) for model predictions, measured in terms of Spearman's $\rho$ correlation over bias scores and F1-score over bias polarity before (rows 1 and 2) and after discounting (rows 3 and 4); and (b) for model representations (rows 5 and 6). * denotes significance at $p<0.001$. 
    %Since 47.2\% of the verbs in the dataset are O-verbs (according to human judgement), this can be considered as a naive baseline in accordance with the models' recency bias.
    }
    \label{tab:corr}
\end{table}

Comparing unidirectional PLMs to bidirectional ones, we find that the latter obtain a stronger correlation to humans scores. A similar trend holds for the F1 scores, where bidirectional models show a greater awareness of the polarity of the IC bias of verbs (especially after discounting). We refrain from making comparisons across model architectures beyond the uni- v. bidirectional dichotomy, to avoid drawing false conclusions: as we are using different procedures to induce a response from generative and discriminative models, it could be argued that a direct comparison is not methodologically robust. The discriminative models are making a binary decision over two options predefined by us, while the generative models are computing a probability distribution over hundreds of thousands of vocabulary items.

\subsection{Further Analysis} To measure the models' sensitivity to IC bias in a perfectly comparable setting, we carry out an additional comparison on the level of representations, thus abstracting away from the top layers of the models where the differences ensue. We extract `decontextualized' verb representation from the PLMs following the procedure described in Appendix~\ref{app:model_reps}. Using those, we carry out two types of probes: an extrinsic one, where we train a linear regression model (LR) to map from a verb's representation to its IC bias; and an intrinsic one, where we use linear discriminant analysis (LDA) to identify the single dimension in the verb representations that is most informative of IC bias.\footnote{As LDA operates over a space of discrete labels, we convert the IC bias scores into 3 classes 
($>0$, $<0$, $=0$).} The benefit of the latter approach is that it does not add any newly trained parameters to the computation of the correlation \cite{torroba-hennigen-etal-2020-intrinsic}. In both cases, the result is a vector of scalars (the values predicted by the LR, or the values of the selected dimension)---we measure the correlation between these values and human IC bias to determine how much of the latter can be recovered from the representations. 

To reduce overfitting, which is inevitable with 305 datapoints in total and representations of 768 to 1024 dimensions, we apply PCA to the representations prior to fitting the LR and LDA models, reducing the representations to 5\% of their original size. Each model (LR and LDA) is fit on a random 50\% split of the data and applied on the other 50\% to predict (LR) or transform (LDA). This procedure is repeated 100 times for robustness. The mean correlations are reported in the last two rows of Table~\ref{tab:corr}. We see that larger models yield lower correlations than their smaller counterparts, suggesting that the former might encode IC bias in a more distributed manner than the latter. 

\begin{figure}[tb!]
    %    \begin{subfigure}{\linewidth}
    %\centering
    %\includegraphics[width=\linewidth]{sections/images-wugs/NP1-ratio-de.pdf}
    %\end{subfigure}
%
 %   \begin{subfigure}[tb!]{\linewidth}
  %  \centering
   % \includegraphics[width=\linewidth]{sections/images-wugs/NP1-ratio-es.pdf}
    %\end{subfigure}
    \includegraphics[width=\linewidth]{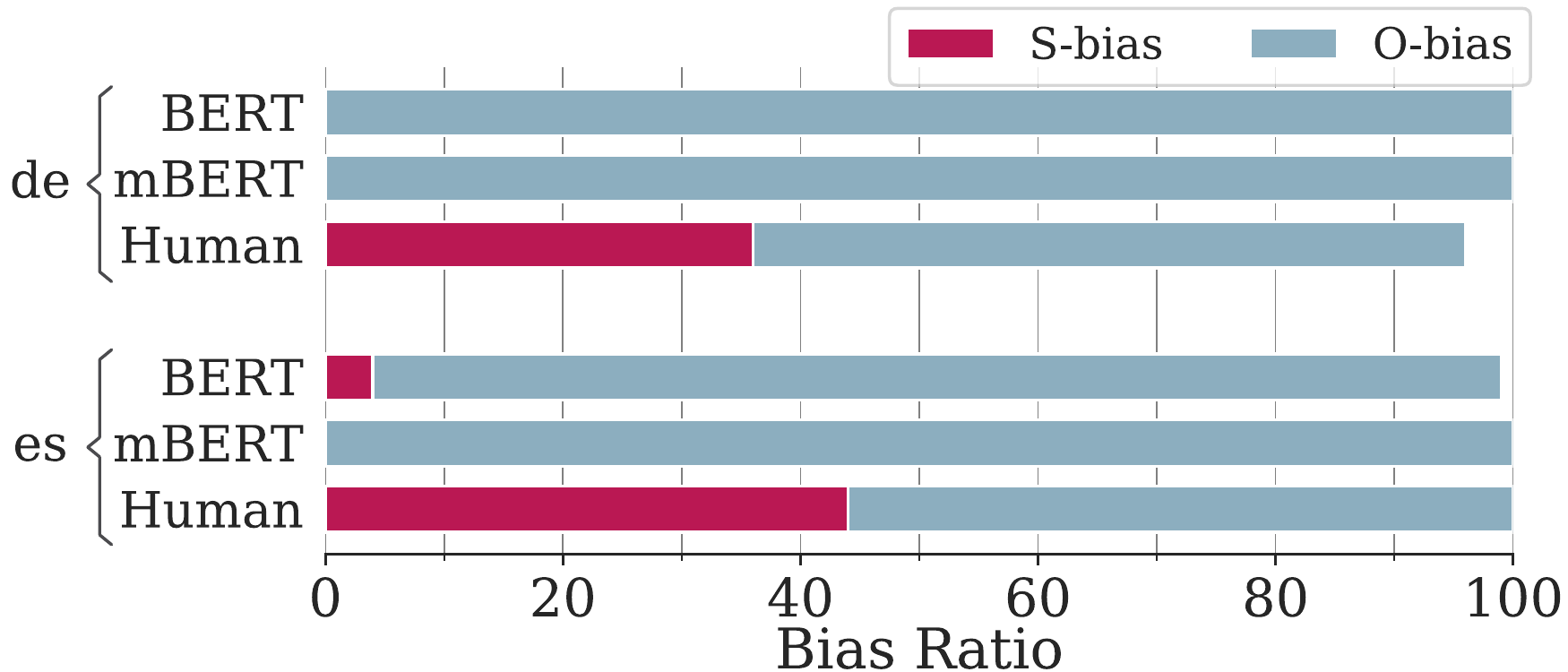}
    \caption{Ratio of S-bias and O-bias verbs in German (top) and Spanish (bottom).}
    \label{fig:ratios_esde}

\end{figure}

Comparing models of equal size, we see that a similar pattern holds here as observed over the models' predictions, with the unidirectional models showing a lower correlation than the bidirectional ones, although the gap is substantially smaller in this space. It appears that unidirectional models might encode more IC bias than they exhibit at inference time. \electra shows the highest correlation among the base-size models and \electralarge the highest among the large models--as this comparison abstracts away from the specific objective each models uses (generative v. discriminative), we can conclude that the two \electra models capture IC bias to the greatest extent out of the six PLMs studied here.

\subsection{IC Bias in Other Languages}

IC bias is not an English-specific phenomenon---\citet{goikoetxea2008normative} obtained human judgements for 100 Spanish verbs, and \citet{van2018discourse} did so for 100 German verbs. %\footnote{We obtained the German data in personal communication with the Emiel.} 
Here, we probe Spanish (es) \bert, German (de) \bert and \mbert for their IC bias awareness.  Details on the choice of proper nouns and nonce words are discussed in Appendix~\ref{app:esde}. As seen in Figure~\ref{fig:ratios_esde} a recency/object bias is observed for the PLMs investigated here as well, so we present the results with and without discounting.

Table~\ref{tab:corr_de_es} summarizes the results before (rows 1 and 2) and after discounting (rows 3 and 4). The poor performance of the multilingual \mbert is not surprising---\citet{ronnqvist-etal-2019-multilingual} found \mbert to be inferior to monolingual models at making a prediction for randomly masked subtokens (specifically looking at German, among other languages); and \citet{vulic2020probing} found \mbert and \xlm to both be inferior to their monolingual %ly-trained 
counterparts on probing tasks pertaining to lexical semantics. 

German \bert shows a medium-strength correlation to human scores, whereas Spanish \bert shows no such correlation at all, both on the level of predictions and model representations. This observation could be attributed to the pro-drop nature of Spanish, wherein pronouns are often dropped when in subject position. This likely makes the learning of IC bias in Spanish harder for a PLM, as less evidence is available in the context to connect the explanation to its referent. 

\begin{table}[t!]
    \centering
    \resizebox{0.8\linewidth}{!}{
    \begin{tabular}{l|lllll}
         \toprule
        &\multicolumn{2}{c}{German} &\multicolumn{2}{c}{Spanish}\\ 
        &\bert &\mbert &\bert &\mbert\\
        \midrule
        
        $\rho$  &  0.54* &  0.23 & 0.13  & -0.00\\
        F1 & 0.600  & 0.600 & 0.540 & 0.560 \\
        \midrule
        $\rho$  &  0.51* &  -0.13  & 0.16  & -0.15\\
        F1 & 0.680  & 0.380 & 0.610 & 0.360 \\
        \midrule
        \midrule
        LDA & 0.26* & -0.0 &  0.09 & 0.08\\
        LR & 0.47* & 0.02 & 0.12 & 0.03\\
    \bottomrule
    \end{tabular}
    }
    \caption{Correspondence between human- and model-induced IC bias in German and Spanish. For more details see the caption of Table~\ref{tab:corr}.
    }
    \label{tab:corr_de_es}
\end{table}

From this section, we conclude that English bidirectional PLMs reliably capture and use IC bias in their predictions. Unidirectional models encode IC bias but do not greatly rely on it at prediction time. Having established that IC bias affects the behavior of at least some PLMs, we now evaluate how these models integrate this implicit signal with more explicit signals from the sentence-level semantics. 

\section{Exp. 2: IC Bias in the Context of EC}

With this experiment, we test the hypothesis that when the IC  and EC signals converge in congruent contexts, i.e. they point to the same referent, the models would have more ease predicting the correct referent, whereas when the two signals diverge in incongruent contexts, the models would be more prone to errors. We test this hypothesis using the materials described in \S\ref{subsec:exp2} and the procedure from \S\ref{sec:procedure}. We present each stimulus to the models in 200 versions varied for subject and object referents. In this experiment, we do not perform discounting: unlike Experiment 1 where we wanted to gain as clear a view as possible of the level of IC bias that models exhibit, isolated from other sources of bias, here we want to see how IC bias interacts with EC, subject to any other potential sources of bias.

\subsection{Results}

\begin{figure}
    \centering
    \includegraphics[width=\linewidth]{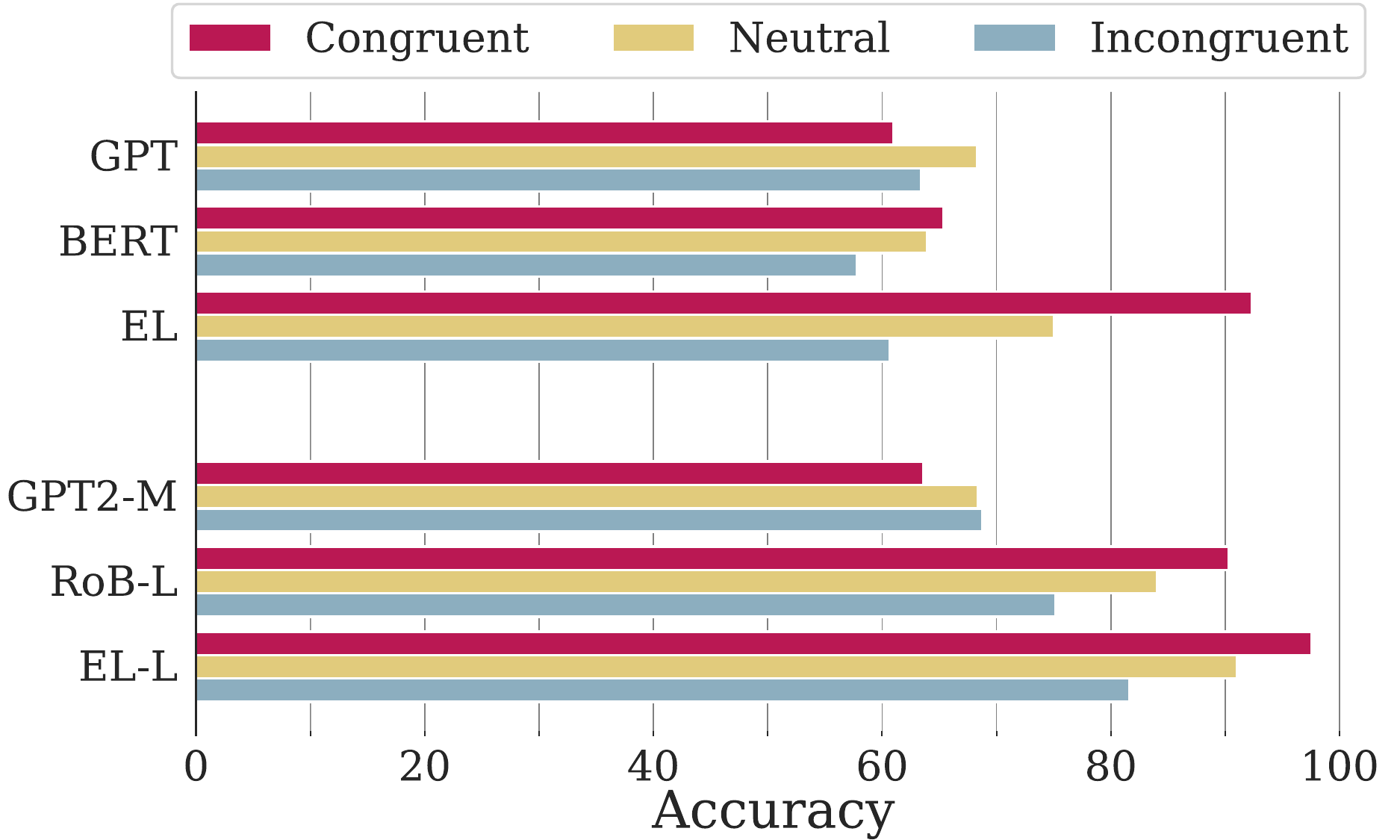}
    \caption{Accuracy on co-reference resolution over stimuli with congruent and incongruent IC and EC.}
    \label{fig:exp2}
\end{figure}

Figure~\ref{fig:exp2} shows the results from this experiment. All models are substantially better at resolving the antecedent correctly on average compared to a random baseline of 50\%. Looking at the neutral stimuli as indication for the models' general ability to solve this task, we see that the two largest and most richly trained models, \roberta and \electralarge, perform best. In line with our expectations, we see that most models score lower on resolving antecedents in incongruent contexts and higher in congruent ones. This is true for the four bidirectional PLMs, which also exhibited higher IC bias in Experiment 1. The gap is largest for \electra and still substantial for \roberta and \electralarge. The unidirectional models, on the other hand, show a noisier behavior, with a relatively small gap between the three types of stimuli, and an inconsistent ranking between them. 

\subsection{Discussion}

As IC bias contributes to the construction of causal links in narrative and as such aids language comprehension \cite{trabasso1985causal}, it is desirable that PLMs capture and use this signal coming from the lexical semantics of interpersonal verbs. In Experiment 1, we found that all PLMs studied show a medium to high correlation with human IC bias scores on the level of representations, with bidirectional ones doing so on the level of predictions, too.  %\citet{ettinger2020bert} argues that probing PLMs in their natural setting (as language models) provides an insight into what kinds of information PLMs truly picked up on during training, presumably as opposed to what kinds of information can be spuriously recovered from their representations \cite{barrett-etal-2019-adversarial,hewitt-liang-2019-designing}. %Furthermore, 
While IC bias does contribute to language comprehension, it also has the unfortunate effect of interference in the context of incongruent EC. In this respect, models with higher IC bias awareness, i.e. the bidirectional PLMs in our study, suffer a greater drop in performance. Meanwhile, the unidirectional PLMs studied, which show little awareness of IC bias in a context free of EC, also show no interference from it when resolving referents in the presence of EC. Paradoxically, the superior performance of bidirectional PLMs with respect to IC bias also exposes a limitation of theirs: while these models are advanced enough to use IC bias for their predictions, their interpretation of semantics is still fairly shallow. The lower-order signal coming from lexical semantics is given priority over the higher-order signal coming from the sentence-level semantics. In the experiment presented in this section, this leads to a higher error rate on resolving pronoun antecedents in incongruent contexts, with potential impact on tasks that depend on co-reference resolution, e.g. document summarization \cite{azzam-etal-1999-using}, question answering \cite{morton-1999-using,vicedo-ferrandez-2000-importance}, and information extraction \cite{zelenko-etal-2004-coreference}.

\section{Conclusion}
From the comparison of six competitive PLMs,  \bert , \roberta , \gpt, \gpttwo, \electra and \electralarge, we conclude that PLMs can exhibit IC bias much like humans do, but that different models do so to a different degree,  with bidirectional models showing moderate to strong correlation to human judgements, and unidirectional models showing only a weak correlation.  This ability of some PLMs has the unfortunate effect that it makes them prone to higher error rates in contexts of incongruent IC and EC signals, where the PLMs overly rely on IC bias. This finding adds to a growing body of evidence that PLMs prioritize lexical cues over higher-order semantic cues (\citealp[cf.][]{tenney2019bert}). As our hypothesis is inspired by the observation that humans experience a delay in the processing of incongruent contexts \cite{caramazza1977comprehension}, our findings point to the potential of drawing further inspiration from such psycholinguistic phenomena in studying the behaviour of language models \cite{ettinger2020bert}. Seeing that language models show a growing potential as off-the-shelf task solvers \cite{radford2019language, brown2020language}, studying their predictions is an important avenue for better understanding their capabilities and limitations.

\section{Acknowledgements}

We thank Daniel Hershcovich, Ana Valeria González, Emanuele Bugliarello, and Mareike Hartmann for feedback on the drafts of this paper. We thank Desmond Elliott, Stella Frank and Dustin Wright, and Mareike Hartmann for their help with the annotation of the newly developed stimuli. 

Yova was funded by Innovation Fund Denmark, under the AutoML4CS project. Mark received funding from the European Research Council (ERC) under the European Union's Horizon 2020 research and innovation programme (FASTPARSE, grant agreement No 714150) and from the Centro de Investigaci{\'o}n de Galicia (CITIC) which is funded by the Xunta de Galicia and the European Union (ERDF - Galicia 2014-2020 Program) by grant ED431G 2019/01.  
\typeout{} 
%\clearpage
\bibliography{biblio}
\bibliographystyle{acl_natbib}

%\afterpage{\null\newpage}
%\newpage
%\clearpage
\appendix

\section{Alternative task formulation}

\subsection{Splitting the clauses}
\label{app:alttask}

\citet{hartshorne2013verb} induce a response by presenting participants with a sentence like 

\begin{enumerate}
   \item[(5)]\label{itm:incon5} John praised Tim because he was a \textit{dax}. 
\end{enumerate}

and asking them the question \textit{Who do you think is a dax?} In the spirit of \citet{radford2018improving} and \citet{radford2019language} we considered reformulating this into a task suitable for a language model as:

\begin{enumerate}
   \item[(6)]\label{itm:incon6} John praised Tim because he was a \textit{dax}. The one who was a \textit{dax} was \underline{\hspace{0.5cm}} 
\end{enumerate}

This formulation would have allowed us to use unidirectional PLMs in a more natural way, for next-word prediction, since the token of interest now comes at the end of the sequence; and it would have been equally suitable for bidirectional PLMs. We performed experiments with it and found that models largely scored at chance level for the stimuli containing neutral verbs, which renders the results for the congruent and incongruent stimuli invalid. The difficulty PLMs faced in solving this task could stem from the more complex inference required and/or from the border-line unnatural structure of the inputs. 

\subsection{Swapping the clauses}
\label{app:swap}
Seeing that all models show some degree of object bias, we considered an alternative task formulation, where the main clause and the subordinate clause are swapped:

\begin{enumerate}
   \item[(7)]\label{itm:incon7} Because \underline{\hspace{0.5cm}} was a \textit{dax}, John praised Mary. 
\end{enumerate}

\begin{figure}[b!]
    \centering
    \includegraphics[width=\linewidth]{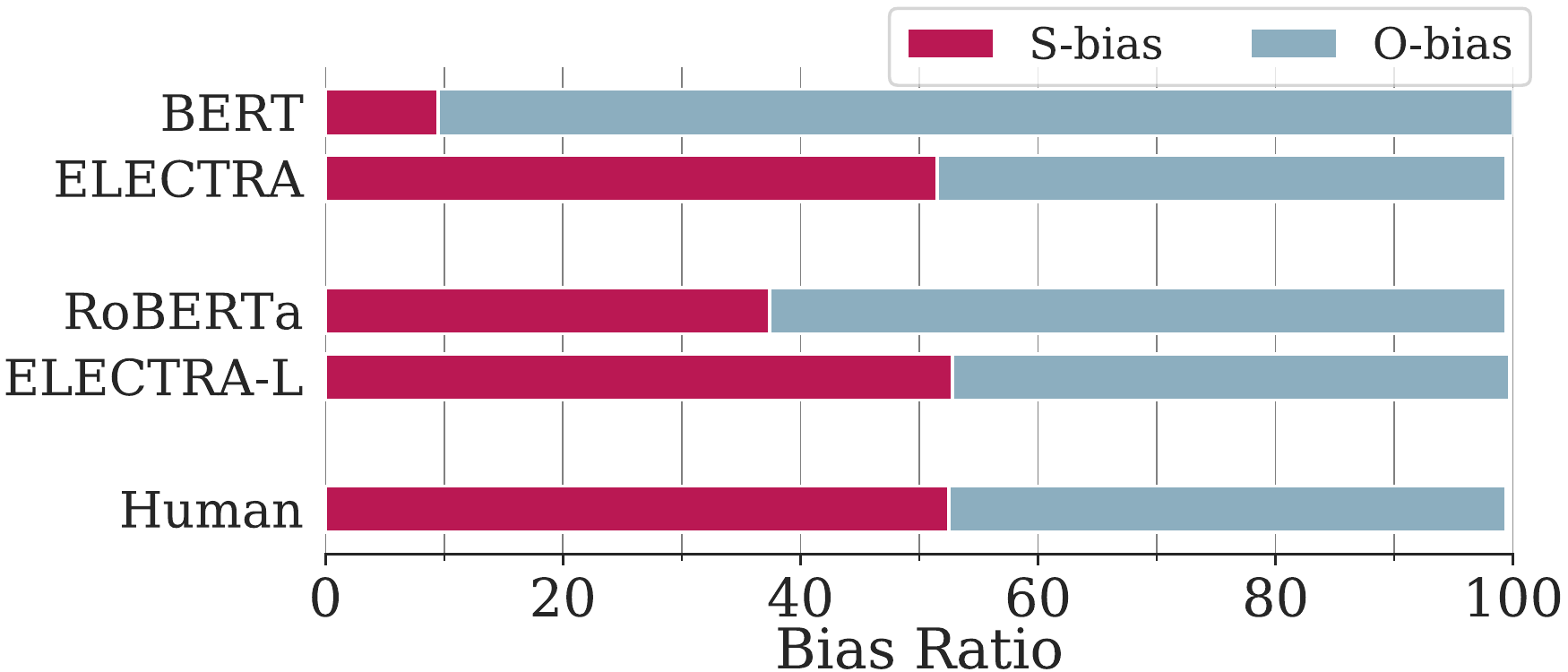}
    \caption{Ratio of S-bias verbs and O-bias verbs with swapped clauses.}
    \label{fig:np1-ratios-new-experiment}
\end{figure}

\noindent In this formulation, the proximity changes for the two referents, such that now the subject is closer to the pronoun of interest. This clause-swapping can only be applied to bidirectional models for the purposes of measuring context-free IC bias as described in Section~\ref{subsec:exp1}. Figure~\ref{fig:np1-ratios-new-experiment} shows the results obtained with this task formulation. Comparing these numbers to the ones presented in Figure~\ref{fig:np1-ratios}, we see that all models show a more balanced distribution of verbs across the S-bias and O-bias classes, with the two \electra models closely matching human scores. This suggests that reference proximity is indeed a factor in the choice of the pronoun. Still, we see that \roberta and especially \bert remain strongly biased towards the object of the sentence, meaning that proximity is not the only factor at play. 

Although this task formulation appears to lead to a reduced object bias across the four bidirectional models, we refrain from using it in our main experiments because it is not attested in psycholinguistic studies, i.e. it could have unforeseeable effects on human judgements of IC bias. As our main goal is to analyse the models’ behavior \textit{in relation to human behavior}, we follow closely the experimental protocol used in the available psycholinguistic studies on IC bias.

\section{Proper nouns}
\label{app:proper}
%We end this section with one last, 
In this section we consider a seemingly minor but important consideration. \citet{abdou-etal-2020-sensitivity} showed that model predictions on the Winograd Schema Challenge greatly vary with changes in the gender and identity of proper nouns used in the stimuli. We alleviate this issue by marginalizing over a range of proper nouns. To do so, we create multiple versions of the same stimulus with different proper noun combinations and use the average response of a model over all of the stimuli as an indication of the model's response to an abstract subject and object. We use 10 male names and 10 female names in 200 permutations. To ensure that the names in these lists are perceived as common names for the gender they represent, we used the models themselves to select the names, compiling a list of names unique to each model. We queried each model with the following sequences and took the top 10 names predicted:\footnote{The mask tag and full-stop were omitted for \gpttwo. As \electra cannot be used in this fashion, we instead used the generative counterpart of the model to obtain the list of names and confirmed that \electra accepts them in their respective contexts (i.e. that it labels them as \textit{original} tokens).} \textit{She is a woman and her name is }{\sc mask}. and \textit{He is a man and his name is} {\sc mask}. %The procedure for German and Spanish is described in Appendix~\ref{app:proper}. 

\section{Development of Materials}
\label{app:materials}

Neither \citet{caramazza1977comprehension} nor \citet{garnham1996locus} provide an explicit description of the procedure used to design their materials, so we extrapolate their methods by observing the materials themselves: the main goal in constructing a stimulus is to ensure that a particular ending is unambiguous (in a standard, most-likely reading) in pointing to exactly one of the two referents. The explanations always start with a verb in the past tense, e.g. \textit{had done well}. And they are simple in the sense that they require little background knowledge. Using these observations as guidelines,  we manually constructed pairs of congruent-incongruent contexts for 99 verbs, i.e. 198 stimuli in total. The materials were validated by three native English speakers and one fluent English speaker, who were asked to perform the cloze task on one stimulus from each pair and to mark ambiguous cases as such rather than making a guess at random. Eight contexts were judged as ambiguous and replaced with a better alternative, also validated in turn. 

\begin{figure*}[tb]
    \centering
    \includegraphics[width=0.92\linewidth]{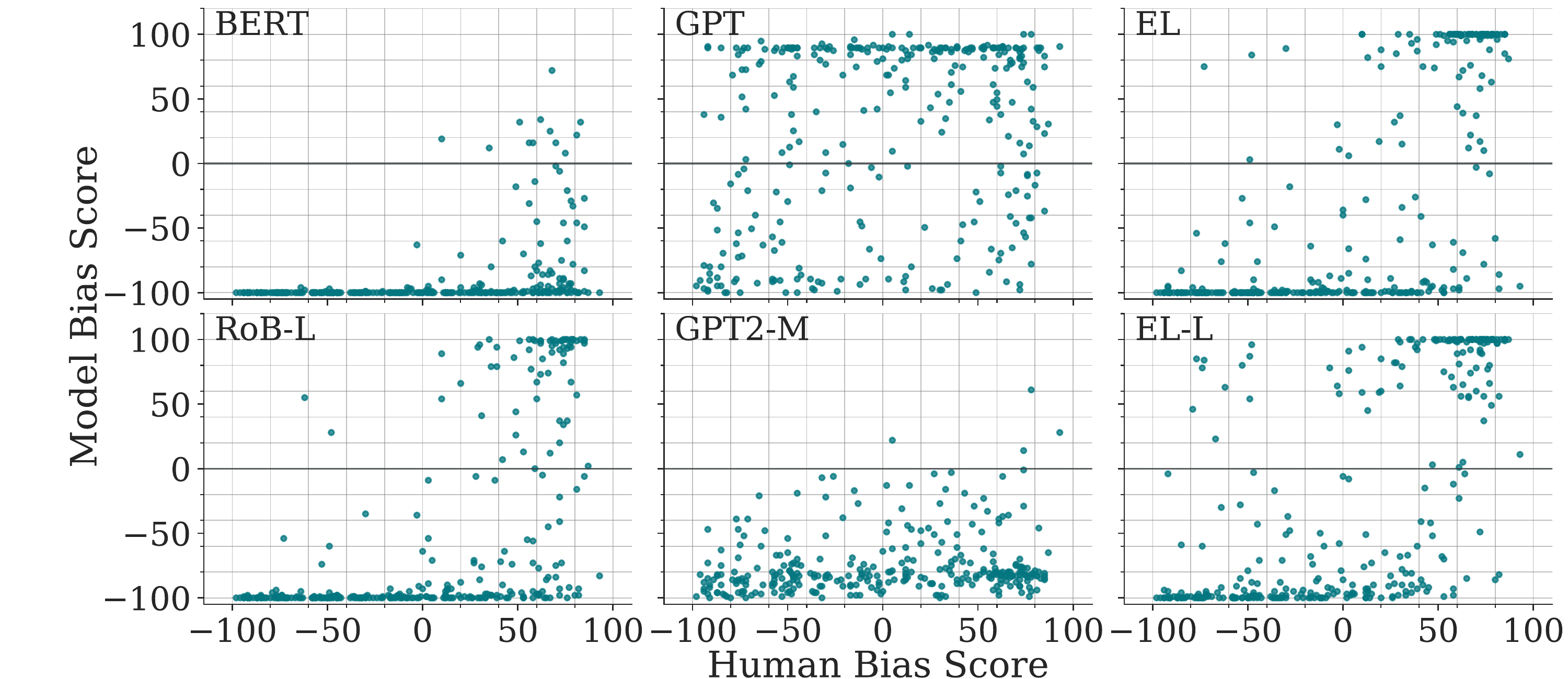}
    \caption{Model bias to human bias agreement.}
    \label{fig:correlations-wugs}
\end{figure*}

\section{Model implementations}
\label{app:models}

We use the models as shared in the HuggingFace library \cite{DBLP:journals/corr/abs-1910-03771}. In English experiments, we use bert-base-uncased, electra-large/base/small-discriminator, gpt2, and roberta-base. For German we use bert-base-german-cased, german-nlp-group/electra-base-german-uncased, dbmdz/german-gpt2, for Spanish dccuchile/bert-base-spanish-wwm-cased, and for both we use xlm-roberta-base and bert-base-multilingual-cased.

All experiments were run on a MacBook Pro, each taking between 1h (for smaller models) to 4h (for larger models).

%\section{Nonce endings: Spanish and German}

%As data equivalent to that used to extract the explanation scaffolds for English is not available for Spanish and German, we used a simple, fixed structure for nonce endings, equivalent to \textit{because X wugged the wug}, i.e. the explanation consists of a pronoun, a verb in the past tense and a noun phrase in the position of the object to this verb. The noun phrase consists of a determiner and singular noun. The order between these elements for Spanish is {\sc pron verb det noun} and for German it is {\sc pron det noun verb}. We sourced nonce verbs in German from \citet{germanverbs}, and in Spanish from \citet{spanishverbs}; We sourced nonce nouns in German from \citet{germannouns1} and \citet{germannouns2}, and in Spanish from \citet{spanishnouns}. 

\section{Probing \electra}
\label{app:electra}
Consider the example sentence \textit{The cake is very delicious} and a ``corrupted'' version of it we might present to \electra: \textit{The shoe is very delicious}. The model could give us a label sequence like \{{\sc orooo}\} for the latter, to indicate that shoes do not belong to the world of delicious things. And we could look at the probability distribution for the second token to quantify the strength of \electra's objection to seeing this token in this position. Alternatively, \electra could resolve the conflict with a label sequence like \{{\sc oooor}\} to indicate that a taste-related adjective is not suitable for describing a shoe. In this case, looking at the probability distribution for the word \textit{shoe} could be rather uninformative. This leads us to conclude that taking the average over the probability of label {\sc o} for all tokens in a sequence is more informative than looking at the probability of this label for a single token. 

\section{Experiment 1 visualisation}\label{app:vis}
Figure \ref{fig:correlations-wugs} shows the data corresponding to the results described in Section \ref{subsec:exp1}. We offer this visualisation for the sake of clarity as correlation coefficients can often be misleading without the accompanying data. By and large, the visualisation of the data corroborates the results discussed in the main body of the paper. 

One observation here is that human bias scores are uniformly distributed between the two extremes (100 and -100), whereas model bias scores tend to be closer to the two endpoints. The scores obtained in the human-based study represent the responses of 100 people to the same stimulus. Deviation from the extremities here show that people differ in their judgements to some degree. The scores obtained in our study represent the responses of a single model to 200 variants of the same stimulus. In this sense, it is not surprising that the model’s scores occupy the two extremities -- this shows the models’ consistency in judging a given verb to be subject or object biasing.

\section{Model representations}
\label{app:model_reps}

To obtain a single representation of any given verb from any given model, we encode a sequence like \textit{John praised Mary} and take the representation for the first subtoken of the verb. With \bert's tokenizer, for example, the first subtoken of a verb amounts to the full verb form 57.1\% of the time, and to the root of the verb 19.3\%; in the remaining 22.6\% the unit is `meaningless.' These numbers vary across models, but in all cases, we are looking at contextualized embeddings, so even `meaningless' subtokens should be a valid proxy to the verb's representation. To abstract away from the exact choice of proper nouns, we repeat this procedure for the 200 name variants and take the element-wise average over all the representations.

\section{Spanish and German experiments}
\label{app:esde}

The contexts we used for extracting proper nouns from the non-English models were \textit{Er ist ein Mann und heißt} and \textit{Sie ist eine Frau und heißt } for German and \textit{Ella es una mujer y se llama}  and \textit{El es un hombre y se llama} for Spanish. For {\sc bert de} we used \textit{Sie heißt} and \textit{Er heißt} instead, as these stimuli more consistently yielded names in the high ranks. 
Since the generator for {\sc electra de} 
%\footnote{We went for the better documented and more downloaded \electra; the other option comes with a generator as well, which we considered for obtaining a list of names, but it did not produce any names for the cues \textit{Sie ist eine Frau und heißt}, \textit{Sie ist eine Frau und sie heißt} or \textit{Sie heißt}.} 
is not publicly available, we used a different procedure to obtain the lists of personal nouns for this model: we queried WikiData for the top 100 male and female given names for people from Germany and then scored these with \electra in the context shown above. We selected the 10 names for each gender that yielded highest probability of an \textit{O} label. 

For the nonce-word slot of the stimuli, we sourced nonce nouns from \citet{germannouns1} and \citet{germannouns2} (German) and \citet{spanishnouns} (Spanish). As these lists contain less than 200 words, here sampling for the nonce-word slot in the 200 variants of a stimulus was done \textit{with} replacement.

%\section{Appendices}
%\subfile{sections/plots_galore}
%\label{sec:appendix}

%\begin{figure}[htpb!]
%    \centering
%    \includegraphics[width=\linewidth]{acl2020-templates/sections/images-wugs/congruency_unamb_name.pdf}
%    \caption{Congruency/incongruency accuracy: unambiguous name.}
%    \label{fig:congruency_unamb_name}
%\end{figure}
%\begin{figure}[htpb!]
%    \centering
%    \includegraphics[width=\linewidth]{acl2020-templates/sections/images-wugs/congruency_amb.pdf}
%    \caption{Congruency/incongruency accuracy: ambiguous}
%    \label{fig:congruency_amb}
%\end{figure}
%\begin{figure}[htpb!]
%    \centering
%    \includegraphics[width=\linewidth]{acl2020-templates/sections/images-wugs/bias_free.pdf}
%    \caption{Bias free unamibiguous contexts.}
%    \label{fig:bias_free}
%\end{figure}

\end{document}